\documentclass[sigconf]{acmart}

\usepackage{booktabs} 
\usepackage{enumitem}
\usepackage[ruled, noend]{algorithm2e}
\usepackage{url}
\usepackage{times} 
\usepackage{lipsum}  
\usepackage{graphicx}
\usepackage{amsmath}
\usepackage{wrapfig}
\usepackage{subfiles}
\usepackage{amsfonts}
\usepackage[acronym]{glossaries}
\usepackage{hyperref}
\usepackage{gensymb}
\usepackage{nicefrac}
\usepackage{float}

\newacronym{ae}{AE}{autoencoder}
\newacronym{rl}{RL}{Reinforcement Learning}
\newacronym{ns}{NS}{Novelty Search}
\newacronym{qd}{QD}{Quality-Diversity}
\newacronym{dmp}{DMP}{Dynamic Movement Primitive}
\newacronym{10d}{10D}{10-dimensional}
\newacronym{cnn}{CNN}{Convolutional Neural Network}
\newacronym{bc}{BC}{Behaviour characterization}
\newacronym{nn}{NN}{Neural Network}
\newacronym{dof}{DoF}{Degrees of Freedom}
\newacronym{es}{ES}{Evolutionary Strategy}
\newacronym{me}{ME}{MAP-Elites}
\newacronym{name}{SERENE}{SparsE Reward Exploration via Novelty search and Emitters}
\newacronym{ea}{EA}{Evolutionary algorithm}

\setcopyright{rightsretained}
\copyrightyear{2021}
\acmYear{2021}
\setcopyright{licensedothergov}\acmConference[GECCO '21]{2021 Genetic and Evolutionary Computation Conference}{July 10--14, 2021}{Lille, France}
\acmBooktitle{2021 Genetic and Evolutionary Computation Conference (GECCO '21), July 10--14, 2021, Lille, France}
\acmPrice{15.00}
\acmDOI{10.1145/3449639.3459314}
\acmISBN{978-1-4503-8350-9/21/07}

\begin{document}

\title{Sparse Reward Exploration via Novelty Search and Emitters}

\author{Giuseppe Paolo}
\affiliation{
\institution{AI Lab, SoftBank Robotics Europe\\ 
Sorbonne Universit\'{e}, CNRS, Institut des Syst\`{e}mes Intelligents et de Robotique, ISIR}
\country{Paris, France}}
\email{giuseppe.paolo@softbankrobotics.com}

\author{Alexandre Coninx}
\affiliation{\institution{Sorbonne Universit\'{e}, CNRS, Institut des Syst\`{e}mes Intelligents et de Robotique, ISIR}
\country{Paris, France}}
\email{alexandre.coninx@sorbonne-universite.fr}

\author{Stephane Doncieux}
\affiliation{\institution{Sorbonne Universit\'{e}, CNRS, Institut des Syst\`{e}mes Intelligents et de Robotique, ISIR}
\country{Paris, France}}
\email{stephane.doncieux@sorbonne-universite.fr}

\author{Alban Laflaqui\`{e}re}
\affiliation{\institution{AI Lab, SoftBank Robotics Europe}
\country{Paris, France}}
\email{alaflaquiere@softbankrobotics.com}


\begin{abstract}
Reward-based optimization algorithms require both exploration, to find rewards, and exploitation, to maximize performance.
The need for efficient exploration is even more significant in sparse reward settings, in which performance feedback is given sparingly, thus rendering it unsuitable for guiding the search process.
In this work, we introduce the SparsE Reward Exploration via Novelty and Emitters (SERENE) algorithm, capable of efficiently exploring a search space, as well as optimizing rewards found in potentially disparate areas.
Contrary to existing emitters-based approaches, SERENE separates the search space exploration and reward exploitation into two alternating processes.
The first process performs exploration through Novelty Search, a divergent search algorithm.
The second one exploits discovered reward areas through emitters, i.e. local instances of population-based optimization algorithms. 
A meta-scheduler allocates a global computational budget by alternating between the two processes, ensuring the discovery and efficient exploitation of disjoint reward areas.
SERENE returns both a collection of diverse solutions covering the search space and a collection of high-performing solutions for each distinct reward area.
We evaluate SERENE on various sparse reward environments and show it compares favorably to existing baselines.
\end{abstract}

\maketitle

\section{Introduction}
Embodied agents 
solve tasks by learning a \emph{policy} dictating how to act in different situations.
This is done by evaluating the agent's performance on the task through a \emph{reward function}.
\begin{figure}[t]
    \centering
    \includegraphics[width=.9\linewidth]{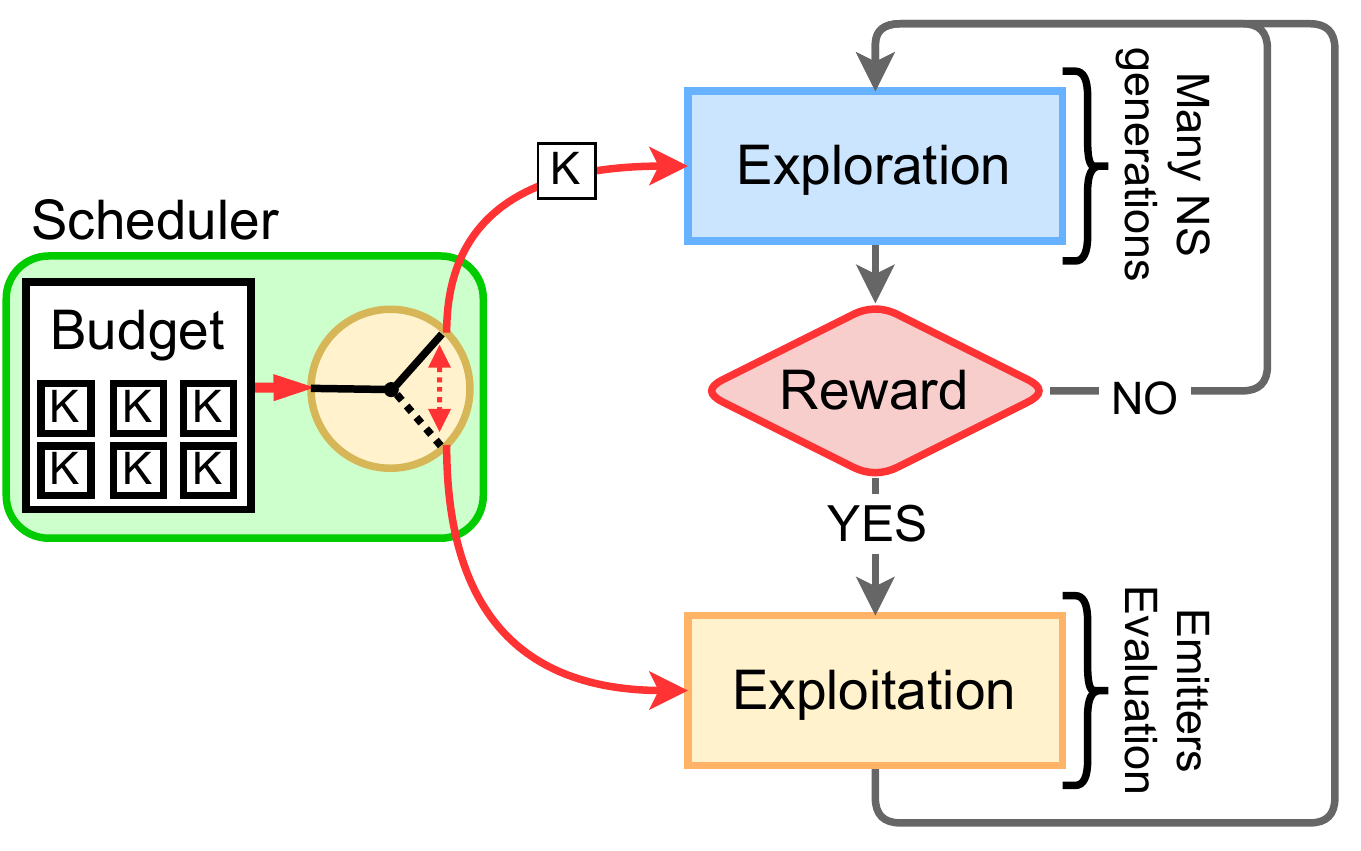}
    \caption{SERENE consists of two exploration and exploitation processes, controlled by a scheduler.
    The exploration process searches for novel solutions through Novelty Search.
    The exploitation process uses emitters to optimize the rewards discovered during exploration.
    The scheduler alternates between the two processes by splitting the total evaluation budget into chunks of size K to assign to either of them.} 
    \label{fig:algo}
\end{figure}
Learning strategies for such agents
can be divided in two groups: step-based and episode-based \cite{sigaud2019policy}.
The former expects a reward after each step.
On the contrary, episode-based ones need rewards only at the end of each training episode.
So much reliance on the reward forces some constraints:
the reward function must be well designed and provide feedback as frequently as possible.
In many complex scenarios where the reward is given only if specific conditions are met, such constraints are impossible to respect.
These are known as \emph{sparse reward} situations and can prove very difficult to tackle.
In this work, we consider sparse reward settings in which the reward is obtained only in small disjoint 
areas of the whole search space.
One example would be a robotic arm trying to push an object to one of a few given positions. 
The search space consists of all the positions the object can achieve, while the reward is given only if the object reaches one of the goals.
In such situations, a standard \gls{rl} \cite{sutton2018reinforcement} agent typically explores by trying random actions.
The probability of finding a reward this way tends to zero, rendering learning impractical.
Therefore, the way \emph{exploration} is performed is fundamental when dealing with sparse rewards settings.

In recent years, many algorithms have been proposed to solve this problem \cite{Ecoffet2019GO_Explore, Lehman2008NS, tang2017exploration, colas2018gep, campos2020explore}.
Among them, \gls{ns} is an evolutionary algorithm that focuses only on exploration, while ignoring any possible reward  \cite{Lehman2008NS}. 
By doing so, \gls{ns} tends towards a uniform exploration of the search space \cite{doncieux2019ns_theory}, avoiding the need for a well-defined reward function.
At the same time, its strength is also its limitation: considering all the non-rewarding areas as valuable as the rewarding ones prevents the algorithm from finding the best possible solutions.
Augmenting \gls{ns} with the ability to shift its focus from pure exploration to reward exploitation could help address this issue. 
One possible way of doing so is by using multi-objective optimization methods like NSGA-II \cite{deb2002fast}.
However, merging exploration and exploitation through a Pareto front can degrade the exploring power of the algorithm.
A different approach is taken by \gls{qd} algorithms, a family of methods that build a set of both diverse and high-quality solutions \cite{pugh2016qdfontier}.

In this work, we introduce \gls{name}, a \gls{qd} algorithm addressing sparse reward problems.
\gls{name} augments \gls{ns} with \emph{emitters} \cite{fontaine2020covariance} to perform rewards maximization while keeping its exploration ability, thanks to a clear separation between the exploration and exploitation.
Introduced as a way to improve the efficiency of \gls{me} \cite{mouret2015illuminating} in the CMA-ME method \cite{fontaine2020covariance}, \emph{emitters} are instances of reward-based evolutionary algorithms scheduled to perform a local search in the search space.
In the original formulation, \gls{me} acts as a scheduler by initializing emitters in different areas of the search space.
The emitters then perform both local exploration and exploitation of the reward, leading to degraded performances in settings with very sparse rewards, where not all policies can obtain a reward.
Conversely, \gls{name} decouples exploration from exploitation to better deal with such situations.
The former is performed through \gls{ns}, completely ignoring the reward.
Once a reward area is found, \gls{name} spawns emitters focusing solely on its maximization, with no dependency on the exploration process.
This allows our algorithm to shift its focus between exploration and exploitation at any moment.
Persisting in exploring even after some reward areas have been found is essential, since other reward areas could be present in the search space.
 
In the following, we will discuss other works tackling the sparse rewards problem in Section \ref{sec:related}.
In Section \ref{sec:background} we will analyze the methods \gls{name} draws from and explain in detail the concept of \emph{emitter}. 
The method itself will be introduced in Section \ref{sec:method}, tested in Section \ref{sec:exps}, and the results discussed in Section \ref{sec:discussion}.
We will conclude with Section \ref{sec:conclusion} by pointing at possible extensions and improvements.
\section{Related work}
\label{sec:related}
\subsection{Sparse rewards}
Step-based algorithms expect a reward at every step, making dealing with sparse reward particularly difficult; this is the case for many \gls{rl} algorithms.
Following the recently increased interest in the problem, many new approaches have been proposed to deal with this sparsity.
Some methods work on improving the data efficiency of the search \cite{Andrychowicz2017HER, Nair2018ImaginedGoal}.
Others introduce some artificial curiosity by counting the number of times a state is visited, and push exploration by making less-visited states more rewarding \cite{bellemare2016unifying, tang2017exploration}.
Another strategy uses additional shaped rewards to aid in approaching the task \cite{trott2019keeping}. 
A population of \gls{rl} agents can also be used to increase exploration while learning a policy \cite{doan2019attraction, jung2020population, parker2020effective}.
However, none of these methods explicitly separates exploration and exploitation.

Episode-based methods, and more specifically evolutionary algorithms \cite{vikhar2016evolutionary}, are better suited for dealing with sparse reward settings, given the more relaxed dependency on the reward.
For this reason, many works combined evolutionary algorithms with \gls{rl}.
Some works use \gls{es} to collect the data over which a \gls{rl} agent is then trained \cite{pourchot2018cem, khadka2018evolution}.
These approaches take advantage of the exploration of evolution-based methods and the higher data efficiency of \gls{rl}. 

Separating exploration from exploitation has proven useful for overcoming deceptive gradients in sparse reward settings \cite{Ecoffet2019GO_Explore, cideron2020qd, colas2018gep}.
In the work from Colas et al. \cite{colas2018gep}, a reward-agnostic exploration phase is first performed through Goal Exploration Processes \cite{forestier2017intrinsically}; then a \gls{rl} based policy is learned on the collected data.
A similar two-step process is used in GO-Explore \cite{Ecoffet2019GO_Explore} to solve ATARI games.
Conversely, QD-RL \cite{cideron2020qd} separates exploration and exploitation by taking advantage of a \gls{qd} population trained through an actor-critic approach. 
Half of the population is optimized for quality, while the other half is optimized for diversity.



\subsection{Divergent search algorithms}
Divergent search methods, as the one used by Cideron et al. \cite{cideron2020qd}, generate solutions by looking for a set of diverse policies.
This prevents getting stuck in local optima that could limit the performance of the solutions.
One of the first algorithms developed in this direction is \gls{ns} \cite{Lehman2008NS}.
Since, many divergent search algorithms have been developed, using different mechanisms to drive the search: curiosity \cite{stanton2016curiosity}, empowerment \cite{campos2020explore}, surprise \cite{Gravina2016Surprise}, diversity \cite{CUlly2015MAPElites, Eysenbach2018DIAYN, cully2017quality, pugh2016qdfontier}, and novelty \cite{lehman2011evolving}.

\gls{qd} \cite{cully2017quality, pugh2016qdfontier} is a family of divergent search algorithms that searches for a set of diverse solutions while also improving on their quality.
A well-known 
\gls{qd} algorithm is \gls{me} \cite{mouret2015illuminating}, a method that drives the search for novel policies by discretizing the search space into a grid and filling its cells with high-performing solutions.

\gls{qd} algorithms have been extended by combining them with \gls{es} \cite{beyer2002evolution} to increase their efficiency and speed of convergence \cite{conti2018improving, fontaine2020covariance, cully2020multi}. 
Conti et al. \cite{conti2018improving} augment an \gls{es} with \gls{ns}'s novelty objective to look for novel solutions while improving their performances. 
At the same time, the approach followed by Fontaine et al. \cite{fontaine2020covariance}, and then extended by Cully \cite{cully2020multi}, uses \gls{me} as a scheduler for modified instances of CMA-ES \cite{hansen2016cma}, named \emph{emitters}.  
Exploration of the search space and reward exploitation are both performed through emitters.
However, fusing the two aspects can limit performances in sparse reward settings where reward-based algorithms struggle to explore. 

In this work, we take inspiration from CMA-ME \cite{fontaine2020covariance} by combining emitters with \gls{ns} to keep the two aspects, i.e. exploration and exploitation, separated.
This allows our method to avoid the shortcomings of exploring through emitters.
In the next section, we describe in detail how both \gls{ns} and emitters work before detailing the functioning of \gls{name}.
\section{Background}
\label{sec:background}
The notation used in this work is based on the one introduced by Doncieux et al. \cite{doncieux2019ns_theory} and is directly inspired by the \gls{rl} literature.

\subsection{Novelty Search}
\label{sec:ns}
\gls{ns} is an evolutionary algorithm that replaces the usual fitness metrics used by evolutionary algorithms with a \emph{novelty} metric.
This metric pushes the search towards novel areas of the search space.
The novelty is calculated in a hand-defined \emph{behavior space} $\mathcal{B}$ in which the behavior of each policy $\theta_i \in \Theta$ is represented.
When a policy is evaluated, it traverses a sequence of states $\tau = [s_0, \cdots, s_T]$, where the initial state $s_0$ is constant for every policy.
Traversed states are observed through some sensors generating a sequence of observations $\tau_{\mathcal{O}} = [o_0, \cdots, o_T]$, with $o_t \in \mathcal{O}$.
From the sequence of observations it is possible to extract a representation $b_i \in \mathcal{B}$ of the policy's behavior by using an observer function $O_{\mathcal{B}} : \mathcal{O} \rightarrow \mathcal{B}$.
This whole process can be summarized by introducing a \emph{behavior function} directly mapping a policy $\theta_i$ to its behavior descriptor $b_i$:
\begin{equation}
\phi(\theta_i) = b_i. 
\label{eq:bd}
\end{equation}
Once computed, the behavior descriptors are used to calculate the policies' novelty as: 
\begin{equation}
\label{eq:novelty}
\eta(\theta_i) = \frac{1}{|J|}\sum_{j \in J}\text{dist}(b_i, b_j)  = \frac{1}{|J|}\sum_{j \in J}\text{dist}\big(\phi(\theta_i), \phi(\theta_j)\big),
\end{equation}
where $J$ is the set of indexes of the $k$ policies closest to $\theta_i$ in the behavior space. 

The novelty of the policies is calculated at each generation and used to choose the policies for the next generation.
Moreover, $N_Q$ policies are sampled to be stored into an \emph{archive}, returned as outcome of the algorithm.
This archive is also used to keep track of the already explored areas of the space $\mathcal{B}$.
This is done by choosing the $|J|$ 
closest neighbors used in equation \eqref{eq:novelty} not only from the current population and offspring but also from the archive.
By choosing the most novel policies from the previous generation to compose the population, the search is always pushed towards less explored areas of $\mathcal{B}$. 
Notwithstanding its capacity for exploration, \gls{ns} cannot exploit the rewards potentially found during the search. 
This can lead to low rewarding solutions.


\subsection{Emitters}
\label{sec:emitters}
An emitter \cite{fontaine2020covariance, cully2020multi} is an instance of a reward-based \gls{ea}, such as CMA-ES \cite{hansen2016cma}. 
Its objective is to rapidly examine a small area of the search space while optimizing on the reward.
The CMA-ME algorithm \cite{fontaine2020covariance, cully2020multi} combines emitters with \gls{me} \cite{mouret2015illuminating}, by using the latter as a scheduler for the emitters evaluation.
It works by initializing a population of policies $\theta$
by sampling their parameters from a distribution $\mathcal{N}(\mu, \Sigma)$ and adding them to the \gls{me} archive.
The algorithm then samples one of these policies and uses it to initialize the population of the emitter $\mathcal{E}_i$.
At this point, $\mathcal{E}_i$ is evaluated until a termination criterion is met; e.g. a lack of increase of the reward found.
Moreover, the policies found during the evaluation of the emitter are added to the \gls{me} archive according to \gls{me} addition strategy.
After the termination of $\mathcal{E}_i$, a new emitter is initialized by sampling another policy from the archive.
This is repeated until the whole evaluation budget is depleted.

Different types of algorithms can be used as emitters, changing how the search is performed and how the policies are selected.
This shows the flexibility of the approach.
At the same time, previous works \cite{fontaine2020covariance, cully2020multi} perform exploration through reward-following emitters.
This reduces performances in situations where the reward is very sparse and many of the policies do not get any reward.
Decoupling the exploitation of the reward from the exploration allows to more efficiently deal with sparse rewards settings \cite{colas2018gep}.
\section{Method}
\label{sec:method}
\gls{name} disentangles the exploration of the behavior space $\mathcal{B}$ from the exploitation of the reward through a two-steps process.
In the first phase, called \emph{exploration phase}, $\mathcal{B}$ is explored by performing \gls{ns}.
As per equation \eqref{eq:bd}, the policies $\theta_i$ found during exploration are assigned a behavior descriptor $\phi(\theta_i)$.
A policy obtaining a reward means that its $\phi(\theta_i)$ belongs to the subspace of rewarding behaviors $\mathcal{B}_\text{Rew} \subseteq \mathcal{B}$.
It is in this subspace that the exploitation of the reward happens.
This is done in the second phase, called \emph{exploitation phase}, in which emitters are initialized using the rewarding policies found in $\mathcal{B}_\text{Rew}$ during exploration.
During the exploitation phase the most rewarding policies are stored to be returned as result of the algorithm.
Moreover, particularly novel policies found by the emitters are also stored. 
By launching emitters only in the neighborhoods of the reward areas, \gls{name} keeps the exploitation of the reward separated from the exploration of the search space.
This results in taking the best of both worlds: the exploration power of \gls{ns} and the focused exploitation of reward-based algorithms.

The exploitation and exploration phases are alternated repeatedly through a meta-scheduler.
This scheduler divides a total evaluation budget $Bud$ in smaller chunks of size $K_{Bud}$ 
and assigns them to either one of the two phases.
The whole process is illustrated in Figure \ref{fig:algo} and described in Algorithm \ref{algorithm}.

\begin{algorithm}
\caption{\gls{name}}\label{algorithm}
\textbf{INPUT:} evaluation budget $Bud$, budget chunk size $K_{Bud}$, population size $M$, emitter population size $M_\mathcal{E}$, offspring per policy $m$, mutation parameter $\sigma$, number of policies added to novelty archive $N_Q$\;
\textbf{RESULT:} Novelty archive $\mathcal{A}_\text{Nov}$, rewarding archive $\mathcal{A}_\text{Rew}$\;
$\mathcal{A}_\text{Nov} = \emptyset$;
$\mathcal{A}_\text{Rew} = \emptyset$\;
$\mathcal{Q}_\text{Em} = \emptyset$;
$\mathcal{Q}_\text{Cand\_Nov} = \emptyset$;
$\mathcal{Q}_\text{Cand\_Em} = \emptyset$\;
Sample population $\Gamma_0$\;
Split $Bud$ in chunks of size $K_{Bud}$\;
\While{$Bud$ not depleted}{
    \If{$\Gamma_0$}{
        Evaluate $\theta_i, ~~ \forall \theta_i \in \Gamma_0$\;
    Calculate $b_i = \phi(\theta_i) \in \mathcal{B}, ~~ \forall\theta_i \in \Gamma_0$\;
    }
    \emph{ExplorationPhase} ($K_{Bud}$, $m$, $\sigma$, $\mathcal{A}_\text{Nov}$, $\mathcal{Q}_\text{Cand\_Em}$, $\Gamma_g$,~$N_Q$)\;
    
    \If{\textbf{not} $\mathcal{Q}_\text{Cand\_Em} == \emptyset$ \textbf{or} \textbf{not} $\mathcal{Q}_{Em} == \emptyset$}{
    \emph{ExploitationPhase} ($K_{Bud}$, $\mathcal{Q}_\text{Cand\_Em}$, $\lambda$, $m$, $\mathcal{Q}_\text{Em}$, $\mathcal{A}_\text{Nov}$, $\mathcal{A}_\text{Rew}$, $M_\mathcal{E}$)\;
    }
}
\end{algorithm}

\begin{figure*}
    \centering
    \includegraphics[width=0.7\linewidth]{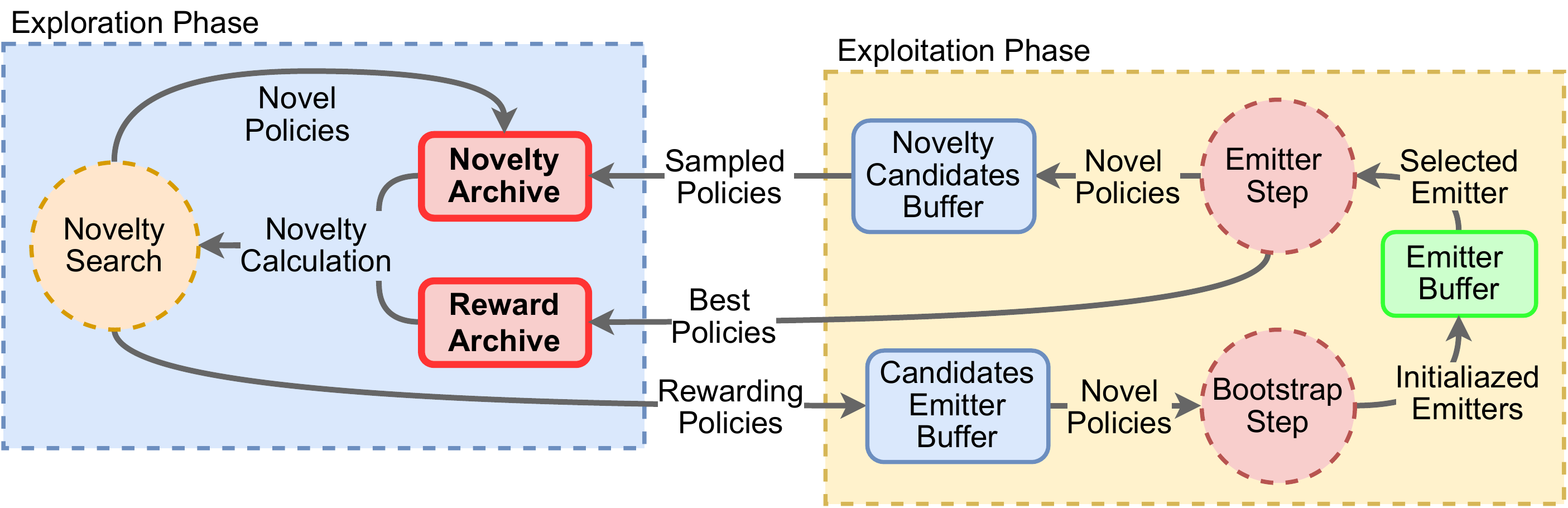}
    \caption{Overview of the sets used by \gls{name} to keep track of the explored areas and the initialized emitters. Highlighted in red are the two archives returned as final result of the algorithm execution. 
    }
    \label{fig:sets}
\end{figure*}

To keep track of policies generated during the different phases, \gls{name} uses the following buffers and containers:
\begin{itemize}
    \item \emph{novelty archive} $\mathcal{A}_\text{Nov}$: a repertoire of the novel policies found during the \emph{exploration phase}, and returned as first output of \gls{name};
    \item \emph{reward archive} $\mathcal{A}_\text{Rew}$: a repertoire of rewarding policies found during the \emph{exploitation phase}, returned as second output of \gls{name};
    \item \emph{candidates emitter buffer} $\mathcal{Q}_\text{Cand\_Em}$: a buffer containing the rewarding policies $\phi(\theta_i) \in \mathcal{B}_\text{Rew}$ found during the \emph{exploration phase} and used in the \emph{exploitation phase} to initialize emitters;
    \item \emph{emitter buffer} $\mathcal{Q}_\text{Em}$: a buffer containing all the initialized emitters to be evaluated during the \emph{exploitation phase};
    \item \emph{novelty candidates buffer} $\mathcal{Q}_\text{Cand\_Nov}$: a buffer containing the most novel policies found by the emitter. Each emitter has its own instance of this buffer and the policies in it are sampled for addition to the novelty archive $\mathcal{A}_\text{Nov}$ once the emitter is terminated. 
\end{itemize}
A high-level overview of how these sets interact during the two phases is given in Figure \ref{fig:sets}, and a more detailed description is proposed in the two following subsections.

\subsection*{Exploration phase}
\gls{name} starts by generating an initial population $\Gamma_0$ of size $M$.
This is done by sampling the parameters of the population's policies $\theta_{j}$ from a normal distribution $\mathcal{N}(0, I)$.
The population is used to explore the behavior space $\mathcal{B}$ through \gls{ns}.
At each generation $g$, a mutation operator generates $m$ new policies $\theta_j^i$ (offspring) from each of the policies $\theta_j \in \Gamma_g$:
\begin{equation}
\label{eq:mutation}
    \forall j, i \in \{1, \dots, M\} \times \{1, \dots, m\}, \theta^i_j = \theta_j + \epsilon, ~~~ \text{with} ~~~ \epsilon \sim \mathcal{N}(0, \sigma I).
\end{equation}
The resulting offspring population $\Gamma^m_g$, of size $m \times M$, is then evaluated to obtain the behavior descriptors $\phi(\theta_j^i) = b_j^i \in \mathcal{B}$.
The novelty of $\Gamma_g$ and $\Gamma^m_g$ is then calculated using equation \eqref{eq:novelty} and is used to generate the next generation population $\Gamma_{g+1}$ by taking the most novel policies from the current population and the offsprings.
At the same time, $N_Q$ policies among the offsprings are uniformly sampled to be added to the \emph{novelty archive} $\mathcal{A}_\text{Nov}$. 
Finally, all the rewarding policies found are stored in the \emph{candidates emitters buffer} $\mathcal{Q}_\text{Cand\_Em}$.
The process just described is detailed in Algorithm~\ref{alg:exploration}.

The exploration phase is executed for the $K_{Bud}$ evaluation steps in the given budget chunk, where each evaluation step corresponds to one policy evaluation.
Once the chunk is depleted, the scheduler assigns the next chunk to the \emph{exploitation phase} only if $\mathcal{Q}_\text{Cand\_Em} \neq \emptyset$.
On the contrary, another \emph{exploration phase} is performed. 
This means that in the worst case scenario where no reward can be discovered, i.e. $\mathcal{B}_\text{Rew} = \emptyset$, \gls{name} performs exactly like \gls{ns}.
\begin{algorithm}
\caption{Exploration Phase}\label{alg:exploration}
\textbf{INPUT:} budget chunk $K_{Bud}$, number of offspring per parent $m$, mutation parameter $\sigma$, novelty archive $\mathcal{A}_\text{Nov}$, candidate emitters buffer $\mathcal{Q}_\text{Cand\_Em}$, population $\Gamma_g$, number of policies $N_Q$\;
\While{$K_{Bud}$ not depleted}{
    Generate offspring $\Gamma^m_g$ from population $\Gamma_g$\;
    Evaluate $\theta_i, ~~ \forall \theta_i \in \Gamma^m_g$\;
    Calculate $b_i = \phi(\theta_i) \in \mathcal{B}, ~~ \forall\theta_i \in \Gamma^m_g$\;
    Calculate $\eta(\theta_i) = \frac{1}{|J|}\sum_{j \in J}\text{dist}(b_i, b_j), ~~ \forall\theta_i \in \Gamma^m_g \bigcup \Gamma_g$\;
    $\mathcal{A}_\text{Nov} \leftarrow N_Q \text{ samples from } \Gamma^m_g$\; 
    \If{$\phi(\theta_i) \in \mathcal{B}_\text{Rew}$}{
        $\mathcal{Q}_\text{Cand\_Em} \leftarrow \theta_i$
    }
    Generate $\Gamma_{g+1}$ from most novel $\theta_i \in \Gamma^m_g \bigcup \Gamma_g$\;
}
\end{algorithm}
\subsection*{Exploitation phase}
The \emph{exploitation phase} consists of two sub-steps: the \emph{bootstrapping step}, in which the policies in the candidates emitter buffer $\mathcal{Q}_\text{Cand\_Em}$ are used to initialize and bootstrap emitters, 
and the \emph{emitter step}, in which the initialized emitters are evaluated.

\subsubsection*{Bootstrap step}
During this step, emitters are initialized from the rewarding policies $\theta_i$ in the candidates emitter buffer, and their potential for reward improvement evaluated.
This insures that only emitters capable of improving the rewards are considered for full evaluation, reducing wasted evaluation budget.
The policies used to initialize the emitters are selected according to their novelty with respect to the reward archive $\mathcal{A}_\text{Rew}$.
This enables \gls{name} to focus on less explored areas of the rewarding behavior space $\mathcal{B}_\text{Rew}$.
The whole bootstrapping phase lasts $\nicefrac{K_{Bud}}{3}$ evaluations.

As discussed in Section \ref{sec:emitters}, an emitter is an instance of a \emph{reward-based \gls{ea}}. 
Contrary to previous work \cite{fontaine2020covariance, cully2020multi}, in this work we do not use estimation-of-distribution algorithms like CMA-ES \cite{hansen2016cma} because the estimation of the covariance matrix $\Sigma$ is unreliable when the population size is smaller than the dimension of the parameter space $\Theta$.
CMA-ES circumvents the issue by using information from previous generations to calculate $\Sigma$.
While stabilizing $\Sigma$, this also leads to a less efficient use of the evaluation budget.
Hence, in this work we use as emitter an \emph{elitist evolutionary algorithm} that does not require any estimation of distribution.
Conversely, it composes its population with the most rewarding policies from the previous generation's population and offspring, while the offspring are generated according to equation \ref{eq:mutation}.

An emitter $\mathcal{E}_i$ based on this algorithm consists of: a population $P$ containing $M_{\mathcal{E}}$ policies $\Tilde\theta \in \Theta$; a population of offspring $P^m$ of size $m \times M_{\mathcal{E}}$; a generation counter $\gamma$; a tracker for the maximum reward found so far $R_{\gamma}$; an improvement measure $I(\cdot)$; a novelty measure $\eta_i$ equal to the novelty of the policy used to initialize the emitter; and a \emph{novelty candidate buffer} $\mathcal{Q}_\text{Cand\_Nov}$.
The emitter $\mathcal{E}_i$ is initialized from a policy $\theta_i$ in the candidates emitter buffer by sampling its initial population $P_0$ from the distribution $\mathcal{N}(\theta_i, \sigma_iI)$.
To keep the emitter's exploration local and prevent overlapping with the search space of possible nearby emitters, we initialize $\sigma_i$ as:
\begin{equation}
    \sigma_i = \frac{\min_{j}\big(\text{dist}(\theta_i, \theta_j)\big)}{3}, ~~\forall \theta_j \in \Gamma^m_g \cup \Gamma_g.
    \label{eq:sigma}
\end{equation}
This shapes $\mathcal{N}(\theta_i, \sigma_iI)$ such that all other $\theta_j$ are at least 3 standard deviation away from its center.
Once $\mathcal{E}_i$ has been initialized, its potential is evaluated by running it for $\lambda$ generations and calculating its \emph{emitter improvement} $I(\mathcal{E}_i)$.
This improvement is defined as the difference between the average rewards obtained during the most recent and the initial generations of the emitter:
\begin{equation}
\label{eq:improv}
    I(\mathcal{E}_i) = \frac{1}{\lambda M_{\mathcal{E}}} \left( \sum_{\gamma=T-\nicefrac{\lambda}{2}}^{T} \sum_{j=0}^{M_{\mathcal{E}}}r_{(\gamma,j)} - \sum_{\gamma=\gamma_0}^{\nicefrac{\lambda}{2}} \sum_{j=0}^{M_{\mathcal{E}}}r_{(\gamma,j)} \right).
\end{equation}
Here $T$ is the last evaluated generation, $r_{(\gamma,j)}$ is the reward of policy $\Tilde{\theta}_j \in P_\gamma$, and $\gamma_0$ is the generation at which the emitter is at the beginning of the exploitation phase;
it is always $\gamma_0 = 0$ for an emitter in the bootstrap step.
If $I(\mathcal{E}_i) \leq 0$, the chances for the emitter to find better solutions than the initial ones are low, so it is not worth allotting more budget to its evaluation.
On the contrary, $I(\mathcal{E}_i) > 0$ means that the emitter has high potential for improvement.
Thus all the initialized emitters for which $I(\mathcal{E}_i) > 0$ are added to the \emph{emitter buffer} $\mathcal{Q}_\text{Em}$ for further evaluation.

\subsubsection*{Emitter step} 
The initialized emitters in the \emph{emitter buffer} $\mathcal{Q}_\text{Em}$ are run during this step.
It starts by calculating the pareto front between the improvement $I(\mathcal{E}_i)$ and the novelty $\eta(\mathcal{E}_i)$ of each of the emitters $\mathcal{E}_i$ in the emitter buffer.
The emitter to run is then sampled from the front of the \emph{non-dominated} emitters.
Using both the novelty and the fitness to select which emitter to run allows \gls{name} to focus both on the less explored and most promising areas of $\mathcal{B}_\text{Rew}$.

The policies $\Tilde{\theta}_j$ generated by an emitter can be stored either for the reward they achieve or for their novelty.
At every generation $\gamma$ all the policies $\Tilde{\theta}_j$ in the current population with a reward $r(\Tilde{\theta}_j) > R_{\gamma-1}$ are added to the reward archive $\mathcal{A}_\text{Rew}$.
Additionally, the policies $\Tilde{\theta}_j$ with a novelty higher than the emitter novelty $\eta_i$ are stored into the emitter's \emph{novelty candidates buffer} $\mathcal{Q}_\text{Cand\_Nov}$.

The emitter $\mathcal{E}_i$ is run until either the given budget chunk is depleted or a termination condition is met.
In the first case, \gls{name} recalculates $I(\mathcal{E}_i)$ from the beginning of the \emph{emitter phase} and assigns the next budget chunk to the \emph{exploration phase}.
On the contrary, if a termination condition is met, $\mathcal{E}_i$ is discarded and another emitter to evaluate is sampled from the Pareto front.
There can be multiple termination conditions. 
The one used in this work is inspired from the \emph{stagnation criterion} \cite{hansen2016cma}, stopping the emitter when there is no more improvement on the reward.
A detailed definition of the termination condition is presented in Appendix \ref{sec:appb}.
Before starting the new emitter evaluation, $N_Q$ policies from the terminated emitter's \emph{novelty candidates buffer} are uniformly sampled to be added to $\mathcal{A}_\text{Nov}$.
In addition to saving particularly novel solutions as part of the final result, this prevents the exploration phase from re-exploring areas covered by emitters during the exploitation phase.

The whole \emph{exploitation phase} is detailed in Algorithm \ref{alg:emitter}.

The code repository is available at: \url{github.com/GPaolo/SERENE}.

\begin{algorithm}
\caption{Exploitation Phase}\label{alg:emitter}
\textbf{INPUT:} budget chunk $K_{Bud}$, candidate emitters buffer $\mathcal{Q}_\text{Cand\_Em}$, number of bootstrap generations $\lambda$, emitter population size $M_\mathcal{E}$, number of offspring per policy $m$, emitters buffer $\mathcal{Q}_\text{Em}$, rewarding archive $\mathcal{A}_\text{Rew}$, novelty archive $\mathcal{A}_\text{Nov}$\;
\CommentSty{*/Bootstrap step/*}\\
\While{$\nicefrac{K_{Bud}}{3}$ not depleted}{
Select most novel policy $\theta_i$ from $\mathcal{Q}_\text{Cand\_Em}$\;
Calculate $\sigma_i$\;
Initialize: $\mathcal{E}_i$, $\mathcal{Q}^i_\text{Cand\_Nov} = \emptyset$, and $P_0$\;
\For {$\gamma \in \{0, \dots, \lambda\}$}{
    \If{$P_0$}{
        Evaluate $\Tilde{\theta}_j$, $\forall \Tilde{\theta}_j \in P_0$\;
    }
    Generate offspring population $P^m_\gamma$ from $P_\gamma$\;
    Evaluate $\Tilde{\theta}_j$, $\forall \Tilde{\theta}_j \in P^m_\gamma$\;
    Generate $P_{\gamma+1}$ from best $\Tilde{\theta}_j \in P^m_\gamma \bigcup P_\gamma$\;
}
Calculate $I(\mathcal{E}_i)$\;
\If{$I(\mathcal{E}_i) > 0$}{
    $\mathcal{Q}_\text{Em} \leftarrow \mathcal{E}_i$\;
}
}
\CommentSty{*/Emitters step/*}\\
Calculate pareto fronts in $\mathcal{Q}_\text{Em}$\;
\While{$\nicefrac{2}{3}K_{Bud}$ not depleted}{
Sample $\mathcal{E}_i$ from \emph{non-dominated emitters} in $\mathcal{Q}_\text{Em}$\;

\While{\textbf{not} $terminate(\mathcal{E}_i)$}{
Generate offspring population $P^m_\gamma$ from $P_\gamma$\;

Evaluate $\Tilde{\theta}_j$, $\forall \Tilde{\theta}_j \in P^m_\gamma$\;

$\mathcal{A}_\text{Rew} \leftarrow \Tilde{\theta}_j, ~~ \forall \Tilde{\theta}_j \in P^m_\gamma \mid r(\Tilde{\theta}_j) > R_\gamma$\; 
$\mathcal{Q}^i_\text{Cand\_Nov} \leftarrow \Tilde{\theta}_j, ~~ \forall \Tilde{\theta}_j \in P^m_g \mid \eta(\Tilde{\theta}_j) > \eta_i$\;

Generate $P_{\gamma+1}$ from best $\Tilde{\theta}_j \in P^m_{\gamma} \bigcup P_\gamma$\;
Update $I(\mathcal{E}_i)$ and $R_\gamma$\;

\If{$terminate(\mathcal{E}_i)$}{
$\mathcal{A}_\text{Nov} \leftarrow N_Q \text{ samples from } \mathcal{Q}^i_\text{Cand\_Nov}$\;
Discard emitter $\mathcal{E}_i$\;
}
}
}
\end{algorithm}
\section{Experiments}
\label{sec:exps}
\begin{figure}
    \includegraphics[width=0.8\linewidth]{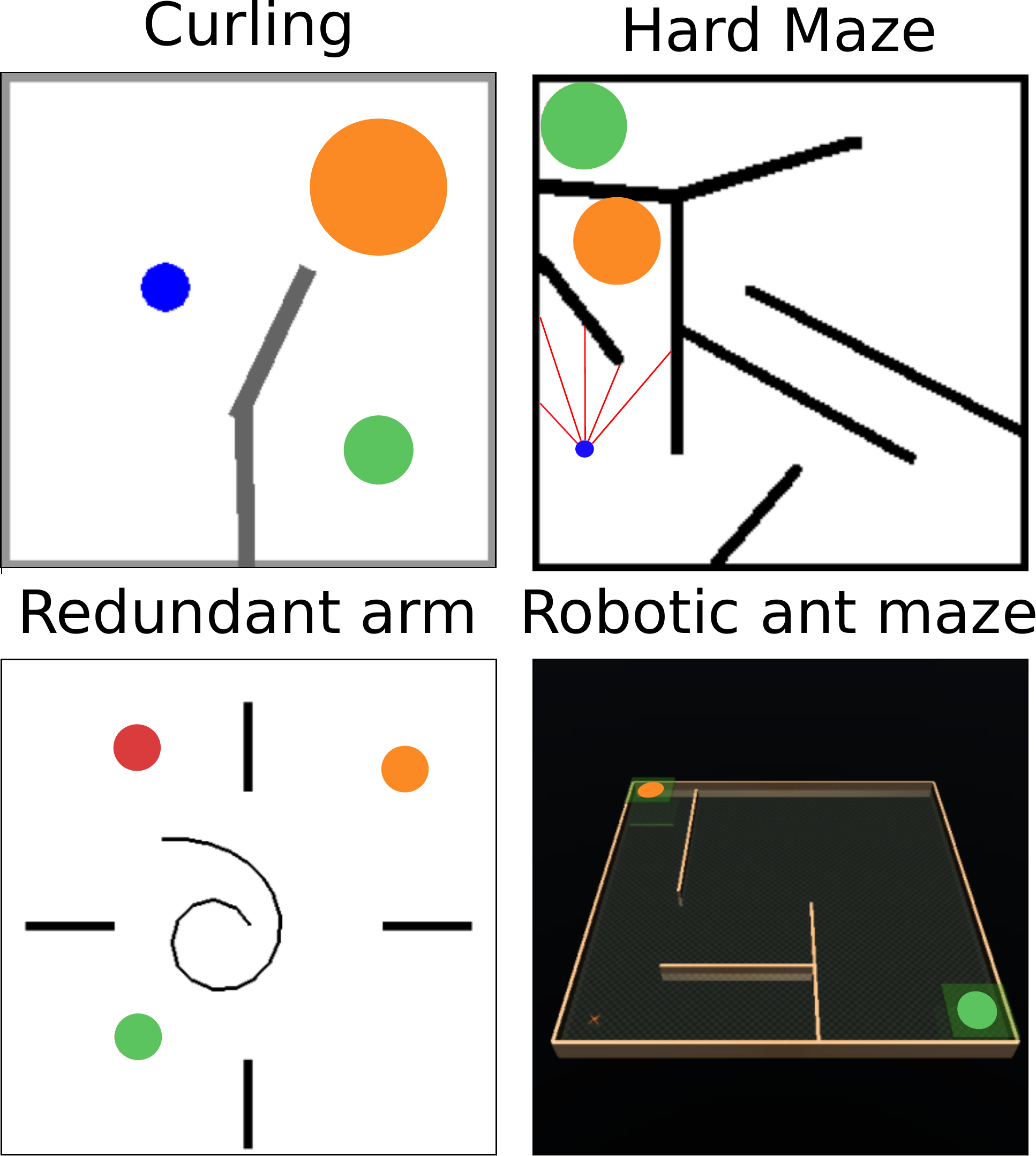}
    \caption{Testing environments: Curling, HardMaze, Redundant arm, Robotic ant maze. 
    }
    \label{fig:environments}
\end{figure}

In this section we want to verify if \gls{name} can efficiently deal with sparse reward settings, find all disjoint reward areas, and optimize the reward in each of them.
For the evaluation, we consider the four sparse rewards environment illustrated in Figure~\ref{fig:environments}:

\noindent\textbf{Curling}: A two \gls{dof} robotic arm controlled by a 3 layers \gls{nn} with each layer  of size $5$. 
The arm has to push the blue ball into one of the two goal areas shown in orange and green. 
A reward is provided only if the ball stops in one of the two areas. 
The controller takes as input a 6-dimensional vector containing the ball pose $(x,y)$, and the two joints angles and velocities. 
The output of the controller is the speed of each joint at the next timestep. 
The size of the parameter space $\Theta$ is 94, and each policy is run in the environment for $500$ timesteps.

\noindent\textbf{Hardmaze}: Introduced in the original \gls{ns} paper \cite{Lehman2008NS}, it consists of a two-wheeled robot, in blue, whose task is to navigate the maze and reach either one of the green and orange areas. 
Contrary to the original formulation, in which only a single binary-reward area was present, here the reward areas are two and provide continuous rewards. At the same time, the reward is only given if the robot stops in one of the two areas.
The robot is controlled by a 2-layers \gls{nn} with each layer of size $5$.
The controller takes as input the reading of the 5 distance sensors mounted on the robot; shown in red in Figure \ref{fig:environments}.
Its output is the 2-dimensional vector containing the speed of the 2 wheels at the next timestep. 
The size of the parameter space $\Theta$ is 63, and each policy is run in the environment for $2000$ timesteps.

\noindent\textbf{Redundant arm}: A 20-\gls{dof} robotic arm \cite{loviken2017online} in which the arm's end-effector has to reach one of the 3 colored goal areas. 
The arm is controller by a \gls{nn} with 2 layers of size 5. 
The controller takes as input the 20-dimensional vector of each joint's position, and outputs the 20-dimensional joint's torque vector. 
The size of the parameter space $\Theta$ is 228, and each policy is run in the environment for $100$ timesteps.

\noindent\textbf{Robotic ant maze}: Introduced by Cideron et al. \cite{cideron2020qd}, it consists in a 4-legged robotic ant in a maze. 
There are two goal areas and the task is for the ant to navigate the maze and reach the center of one of them.
The robot is controlled by a 3-layers \gls{nn}, with each layer of size 10. 
The input of the controller is the 29-dimensional observation returned by the environment at each step, while its output is the 8-dimensional joint's torque control.
The size of the parameter space $\Theta$ is 574, and each policy is run in the environment for $3000$ timesteps.

For all environments, the reward is given only if inside the reward area, and as a continual value in the $[0, 1]$ range. The reward varies with the distance to the center of the area and is highest directly at the center.

It can be expressed as:
\begin{equation*}
    r(\theta) = \begin{cases} 0, & \mbox{if } d_r > radius \\ \frac{radius - d_r}{radius}, & \mbox{if } d_r \leq radius \end{cases}
\end{equation*}
where $d_r$ is the distance from the center and $radius$ is the radius of the reward area.

\subsection*{Baselines}
We compare \gls{name} against 5 different baselines:
\begin{itemize}
    \item \textbf{\gls{ns}}\cite{Lehman2008NS}: vanilla \gls{ns}, that performs pure exploration and does not attempt to improve on the reward;
    \item \textbf{NSGA-II}\cite{deb2002fast}: a multi-objective evolutionary algorithm optimizing both the novelty and the reward; 
    \item \textbf{CMA-ME}\cite{fontaine2020covariance}: the original algorithm introducing emitters that combines \gls{me} with emitters over a $50 \times 50$ grid covering the behavior space of all environments. 
    Among the various emitters proposed by the authors we selected the ``optimizing'' emitter;
    \item \textbf{ME}\cite{mouret2015illuminating}: vanilla MAP-Elites that uses a $50 \times 50$ grid to cover the behavior space of every environment;
    \item \textbf{RND}: pure random search in which no selection happens, and every policy is sampled from a normal distribution $\mathcal{N}(0, I)$.
\end{itemize}
The parameters used during the experiments are listed in Appendix \ref{sec:hyperpar}.
The statistical results are computed over 15 runs for each experiment.
\section{Results}
\label{sec:discussion}
This section discusses the results obtained during the experiments.

%
\subsection{Budgeting}
\label{sec:budget}
Balancing the exploration of the search space and the exploitation of the reward is an aspect of paramount importance for reward-based algorithms.
Even more so in sparse reward environments.
This balance can be studied by analyzing the amount of evaluation budget dedicated to either one of the two aspects.
The exploration budget consists of all the evaluated policies that did not get any reward.
On the contrary, the exploitation budget is obtained by counting all the evaluated policies that collected some reward from one of the reward areas.

\begin{figure}[!ht]
    \centering
    \includegraphics[width=\linewidth]{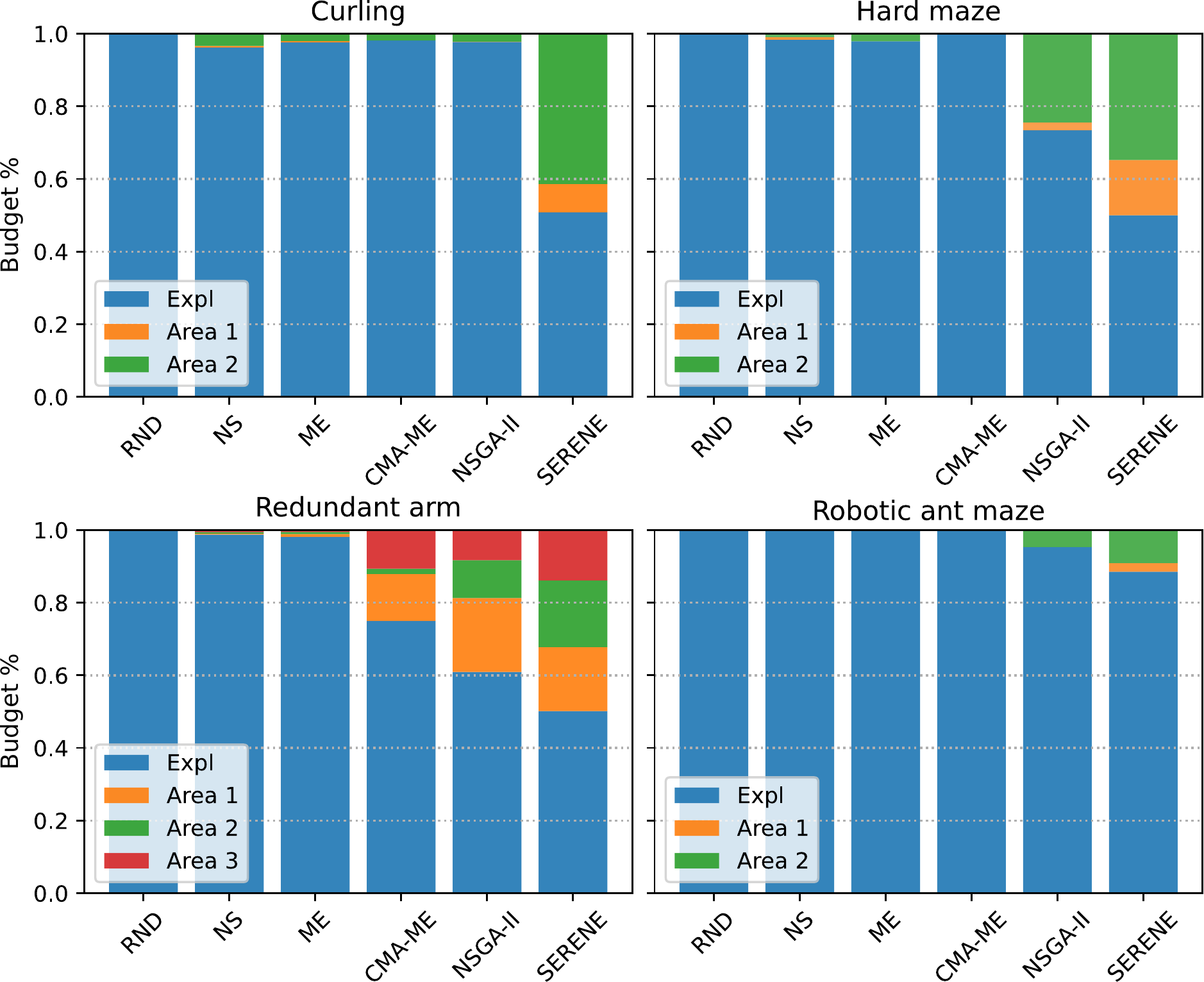}
    \caption{Average budget percentage between the exploration of the search space (in blue) and the exploitation of each reward areas (other colors). 
    }
    \label{fig:focus}
\end{figure}
As Figure \ref{fig:focus} shows, \gls{name} has a more balanced budget split between exploration (in blue) and exploitation (other colors) compared to the other baselines.
In situations in which exploration is harder, a bigger part of the budget is assigned to exploration rather than exploitation of the reward.
This is the case for the robotic ant maze environment.
Additionally, due to the way emitters are selected, the algorithm can shift its exploitation focus among the different reward areas.
Figure \ref{fig:focus} shows that most of \gls{name}'s exploitation budget is assigned to the green reward area in the Curling, Hard maze and Robotic ant maze environments.
As it can be seen in Figure \ref{fig:environments}, this area is more difficult to discover and to reach with respect to the orange area.
This makes the exploitation of the orange reward area faster, having both the novelty and the improvement go to zero rapidly.
On the contrary, being the green area harder to reach, its novelty will remain higher for longer, making \gls{name} select more emitters focused on it.
The effect can also be seen in Figure \ref{fig:rew}, where the reward for area 1 quickly reaches higher values compared to the one of reward area 2. 
At the same time, in the Redundant arm environment where the 3 reward areas are equally easy to discover and to reach, this effect is less present and the exploitation budget is more evenly split between them. 
The ability to switch its focus is similar to intrinsic motivation based methods \cite{gottlieb2013information, BlaesVlastelicaZhuMartius2019} and allows \gls{name} to reach high rewards in all reward areas.
Other baselines exhibit a less balanced distribution of the evaluation budget, as they do not explicitly separate exploration from exploitation.

\subsection{Exploration}
\label{sec:exploration}
\begin{figure*}[!h]
    \centering
    \includegraphics[width=\linewidth]{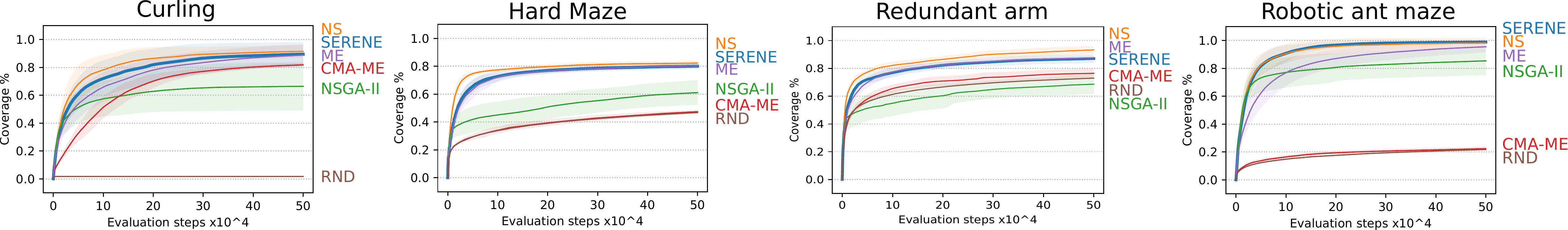}
    \caption{Average coverage with respect to the given evaluation budget. The shaded areas represent one standard deviation.}
    \label{fig:cvg}
\end{figure*}
Performing good exploration in situations of sparse rewards is fundamental in order to discover all the possible rewarding areas of the search space.
In our experiments, we measured the exploration capacity of each of the tested algorithms through the \emph{coverage metric} \cite{mouret2015illuminating, paolo2019unsupervised}.
It is evaluated by discretizing the search space in a $50 \times 50$ grid and calculating the percentage of cells occupied by the policies found during the search.
This metric does not include any measure of the performance of the solutions in the cells.

The plots in Figure \ref{fig:cvg} show that \gls{name} can perform exploration with an efficiency comparable to \gls{ns}, notwithstanding the lower budget assigned to exploring the search space.
At the same time, Figure \ref{fig:cvg} shows that the final coverage obtained by \gls{me} is similar to the one of \gls{ns} and \gls{name}.

On the contrary, although based on \gls{me}, CMA-ME results are more variable across all environments, and exhibit lower exploration compared to \gls{me}.
This effect is likely due to the reliance on emitters for exploration, leading to more local exploration in the parameter space $\Theta$.
It can prove useful in environments like Curling or Redundant arm, where a small change in parameters leads to big behavioral changes, increasing the probability of finding a reward.
On the contrary, environments like Hard Maze or Robotic ant maze in which this does not happen can prove more challenging to explore.

At the same time, the exploration performance of NSGA-II is poor.
In the Redundant arm environment, exploration is even lower than the random search baseline.
This result is likely due to the multi-objective approach of optimizing both novelty and reward through Pareto fronts.
Therefore, as soon as a reward area is discovered, the best strategy to improve the front is to focus on the reward because this scales better than the novelty.

\subsection{Exploitation}
\label{sec:exploitation}
\begin{figure*}[!ht]
    \centering
    \includegraphics[width=.98\textwidth]{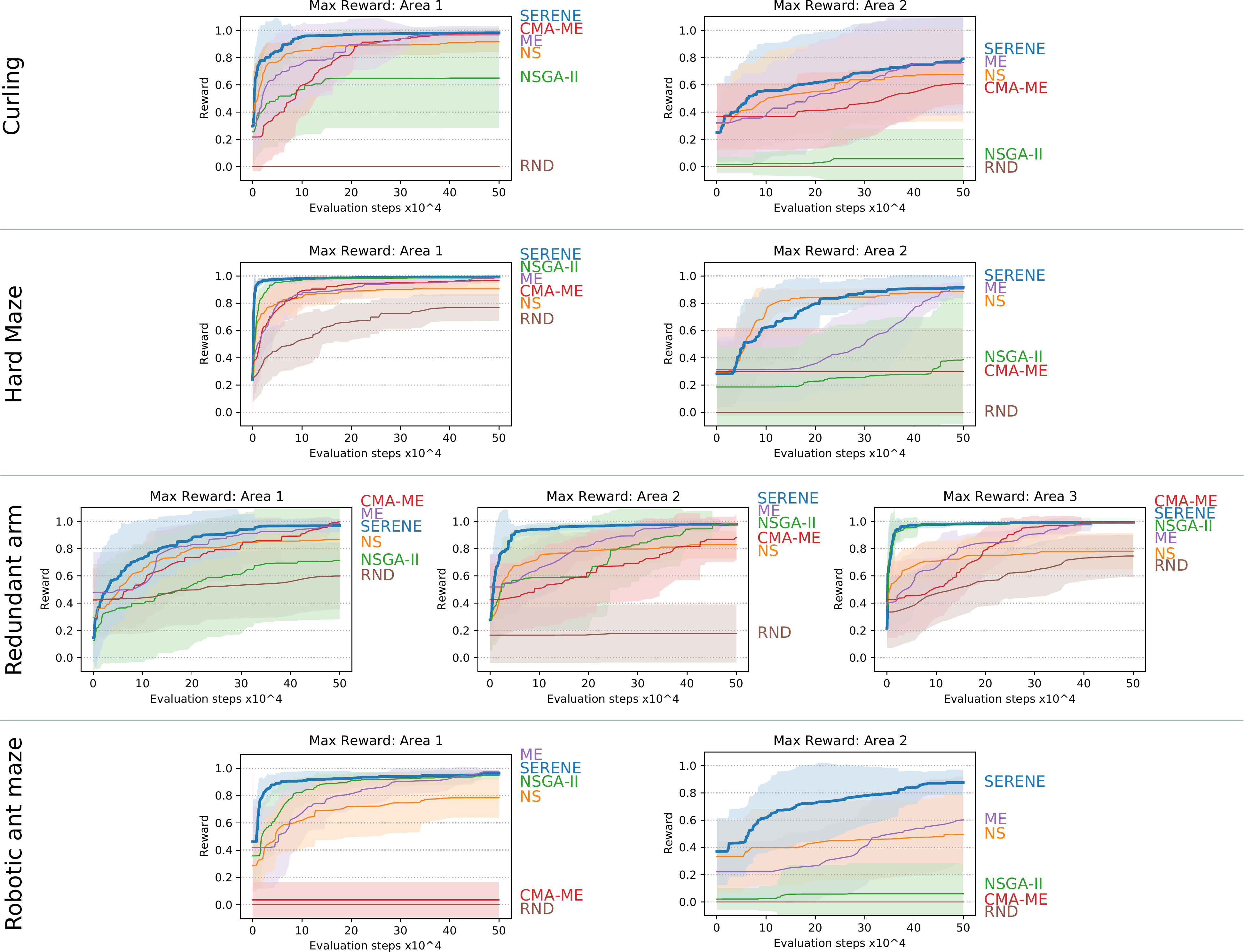}
    \caption{Average maximum reward reached in all the reward areas. The shaded areas represent one standard deviation.}
    \label{fig:rew}
\end{figure*}
Figure \ref{fig:rew} shows the average maximum reward achieved by the algorithms in the reward areas of all environments.
Emitters solely focusing on exploiting the reward allow \gls{name} to reach almost the maximum reward on the easiest to reach reward areas in less than $10^5$ evaluations.
High rewards are also achieved on the harder to reach areas, even if the required time is higher.
On the contrary, \gls{me} improves on the reward at a much slower pace.
This is likely due to the random selection of policies from the archive to generate new policies.
In a sparse reward environment in fact, the probability of selecting a rewarding policy is proportional to the ratio between the rewarding and non-rewarding areas.
The sparser the reward is, i.e. the smaller the reward area is, the lower the probability of selecting a rewarding policy from the archive is, and the slower the exploitation gets. 
A similar trend is exhibited by CMA-ME: even if able to reach high rewards on the discovered reward areas, it is slow in its optimization.
At the same time, even \gls{ns} reached high rewards on almost all environments, but without any explicit reward optimization it did not exploit the reward areas to the maximum.
The multi-objective approach NSGA-II can always find at least one of the multiple reward areas, but then tends to extensively focus on it, instead of also exploring other areas.
For this reason only the easiest reward area is exploited to high values in all environments, while the harder reward area is seldom exploited.
\section{Conclusion and Future Work}
\label{sec:conclusion}
In this work we introduced \gls{name}, a method that efficiently deals with sparse reward environments by augmenting \gls{ns} with emitters.
Contrary to similar methods using emitters, \gls{name} keeps exploration and exploitation of the reward as two distinct processes.
Exploration is carried out by taking advantage of \gls{ns} to discover all the reachable reward areas.
These areas are then exploited by using local instances of population-based optimization algorithms called emitters.
By using a meta-scheduler, \gls{name} can automatically assign the evaluation budget to either exploration or exploitation.
This is advantageous also in situations in which no reward is present: in the absence of reward to exploit, \gls{name} performs exactly like \gls{ns}.

\gls{name} has been tested on four different sparse reward environments, reaching high performances on all of them.
Notwithstanding these encouraging results, the method still suffers from the same limitations as other \gls{qd} methods, and first and foremost from the prior hand-design of the behavior space $\mathcal{B}$.
In the future we will work on addressing this limitation by learning a behavior descriptor that could foster exploration towards rewarding solutions.

At the same time, it has been highlighted by Cully \cite{cully2020multi} that many kind of emitters can be used to address different kind of problems.
Evaluating and combining different types of emitters is also an exciting line of work to extend the current method.


\bibliographystyle{ACM-Reference-Format}
\bibliography{main}
\clearpage
\appendix
\section{Final archive distribution}
\begin{figure*}[!ht]
    \includegraphics[width=.8\textwidth]{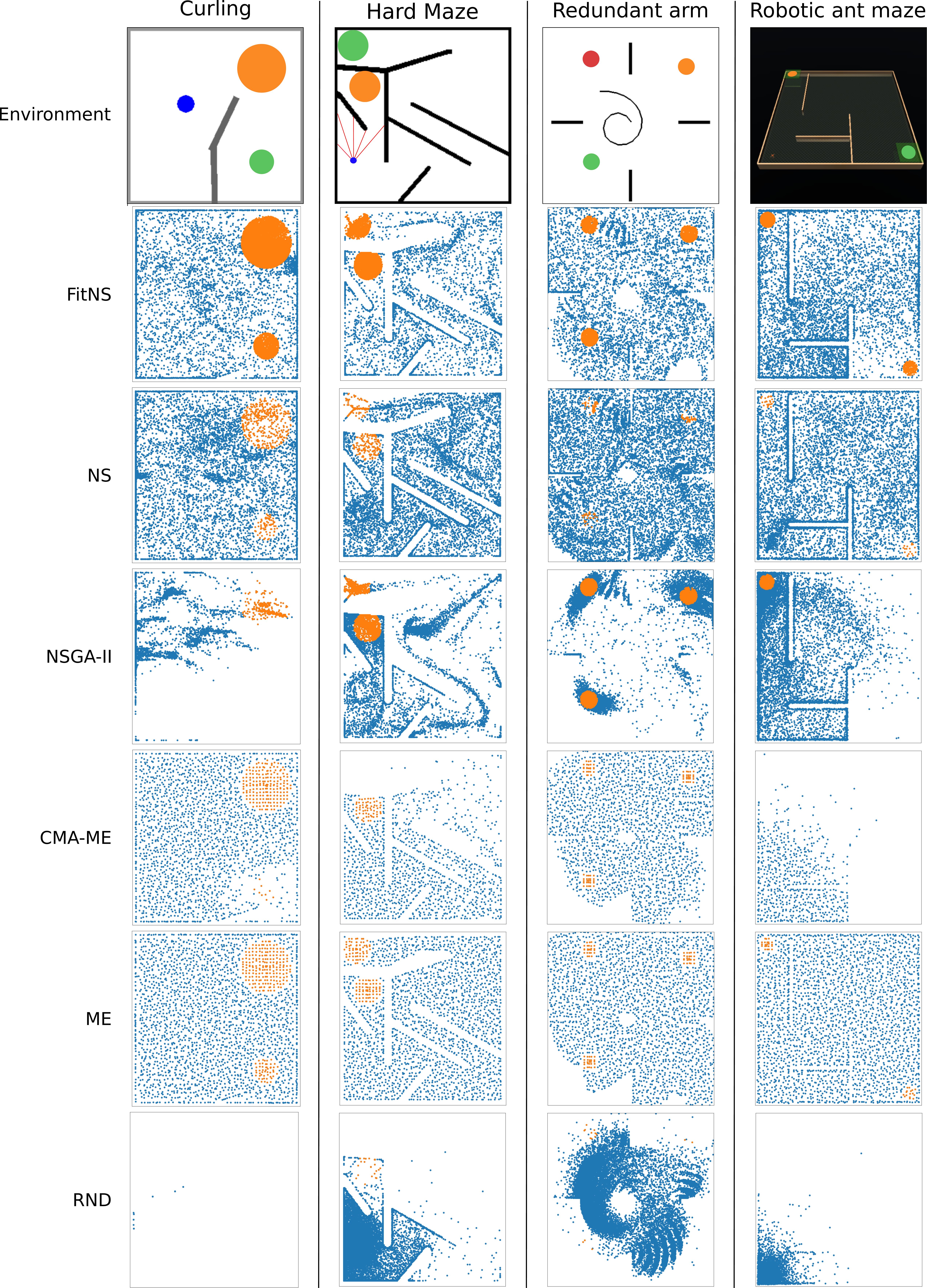}
    \caption{Distribution of the behavior descriptors of the archived policies. On each column are shown the results for an environment, while on each row is shown the distribution for each experiment. The archive plotted are from the runs achieving highest coverage. In blue are the policies with no reward, in orange the policies with a reward. For \gls{name} in blue are the policies in the \emph{novelty archive} and in orange the policies in the \emph{reward archive}. }
    \label{fig:archives}
\end{figure*}

In figure \ref{fig:archives} we show the distribution of the behaviors of the policies in the final archive.
Each point is represents different policy.
In blue are the policies that do not get any reward, thus considered \emph{exploratory}, while in orange are rewarding policies, considered \emph{exploitative}.
For \gls{name} the \emph{exploratory policies} are the ones in the \emph{novelty archive} $\mathcal{A}_N$, while the \emph{exploitative policies} are the ones in the \emph{rewarding archive} $\mathcal{A}_R$.

We can see that even if the coverage metric values for \gls{name} are lower with respect to \gls{me}, the search space is well covered.
Moreover, the reward areas are densely explored.

\section{Hyperparameters}
\label{sec:hyperpar}
The values of the hyperparameters used during the experiments are listed here.
For each experiment we used a budget of $Bud=500000$ evaluations, with the chunk size set to $K_{Bud}=1000$. 
The population size is $M=100$, and for each policy we generate $m=2$ offspring.
As mutation parameter we used $\sigma=0.5$, while the number of policies uniformly sampled to be added to the novelty archive is $N_Q=5$.
\gls{name} uses an emitter population size of $M_\mathcal{E}=6$, with a bootstrap phase for each emitter of $\lambda=6$ generations.
For CMA-ME we used the same parameters used by Fontaine et al. \cite{fontaine2020covariance}: 15 emitters, each one with a population size of 37.
In every experiment, the policies parameters are bounded in the $[-5, 5]$ range.

\section{Termination criterion}
\label{sec:appb}
The termination condition used for our emitters is inspired by the \emph{stagnation criteria} introduced in \cite{hansen2016cma}.
We track the history of the rewards obtained over the last $120 + 20 * n / \lambda$ emitter's generations.
Where $n$ is the size of the parameter space $\Theta$ and $\lambda$ is the emitter's population size.
The emitter is terminated if either the maximum or the median of the last 20 rewards is not better than the maximum or the median of the first 20 rewards.

\end{document}